\documentclass[10pt,twocolumn,letterpaper]{article}

\usepackage{iccv}
\usepackage{times}
\usepackage{epsfig}
\usepackage{graphicx}
\usepackage{amsmath}
\usepackage{amssymb}
\usepackage{epsfig}
\usepackage{graphicx}
\usepackage{amsmath}
\usepackage{amssymb}
\usepackage{booktabs}

\usepackage[pagebackref=true,breaklinks=true,letterpaper=true,colorlinks,bookmarks=false]{hyperref}

\iccvfinalcopy % *** Uncomment this line for the final submission

\def\iccvPaperID{11} % *** Enter the ICCV Paper ID here

\ificcvfinal\fi

\begin{document}

%%%%%%%%% TITLE
%\title{Unsupervised Action Segmentation by Self-Labeling}
%\title{Unsupervised Discovery of Actions for Unannotated Instructional Videos}
\title{Unsupervised Action Segmentation for Instructional Videos}
%\title{Unsupervised Action Segmentation for Instructional Videos}
%\title{Unsupervised Learning of Temporal Structure from Human Example Videos?}

%\author{First Author\\
%Institution1\\
%Institution1 address\\
%{\tt\small firstauthor@i1.org}
% For a paper whose authors are all at the same institution,
% omit the following lines up until the closing ``}''.
% Additional authors and addresses can be added with ``\and'',
% just like the second author.
% To save space, use either the email address or home page, not both
%\and
%Second Author\\
%Institution2\\
%First line of institution2 address\\
%{\tt\small secondauthor@i2.org}
%}

\author{AJ Piergiovanni\\
Robotics at Google\\
%Institution1 address\\
%{\tt\small firstauthor@i1.org}
% For a paper whose authors are all at the same institution,
% omit the following lines up until the closing ``}''.
% Additional authors and addresses can be added with ``\and'',
% just like the second author.
% To save space, use either the email address or home page, not both
\and
Anelia Angelova \\
Robotics at Google\\
%First line of institution2 address\\
%{\tt\small secondauthor@i2.org}
\and
Michael Ryoo\\
Robotics at Google\\
%First line of institution2 address\\
%{\tt\small secondauthor@i2.org}
\and
Irfan Essa\\
Google Research\\
%First line of institution2 address\\
%{\tt\small secondauthor@i2.org}
}

\maketitle

%%%%%%%%% ABSTRACT
\begin{abstract}

In this paper we address the problem of automatically discovering atomic actions in unsupervised manner from instructional videos, which are rarely annotated with atomic actions. We present an unsupervised approach to learn atomic actions of structured human tasks from a variety of instructional videos based on a sequential stochastic autoregressive model for temporal segmentation of videos. This learns to represent and discover the sequential relationship between different atomic actions of the task, and which provides automatic and unsupervised self-labeling.  

\end{abstract}

\section{Introduction}
Instructional videos cover a wide range of tasks: cooking, furniture assembly, repairs, etc. The availability of online instructional videos for almost any task provides a valuable resource for learning, especially in the case of learning robotic tasks. So far, the primary focus of activity recognition has been on supervised classification or detection of discrete actions in videos. However, instructional videos are rarely annotated with atomic action-level instructions.  In this work, we propose a method to learn to segment instructional videos in atomic actions in an unsupervised way, i.e., without any annotations. To do this, we take advantage of the structure in instructional videos: they comprise complex actions which inherently consist of smaller atomic actions with predictable order.  While the temporal structure of activities in instructional videos is strong, there is high variability of the visual appearance of actions, which makes the task, especially in its unsupervised setting, very challenging. For example, videos of preparing a salad can be taken in very different environments, using kitchenware and ingredients of varying appearance.

\begin{figure}
%    \centering
\includegraphics[width=0.8\linewidth]{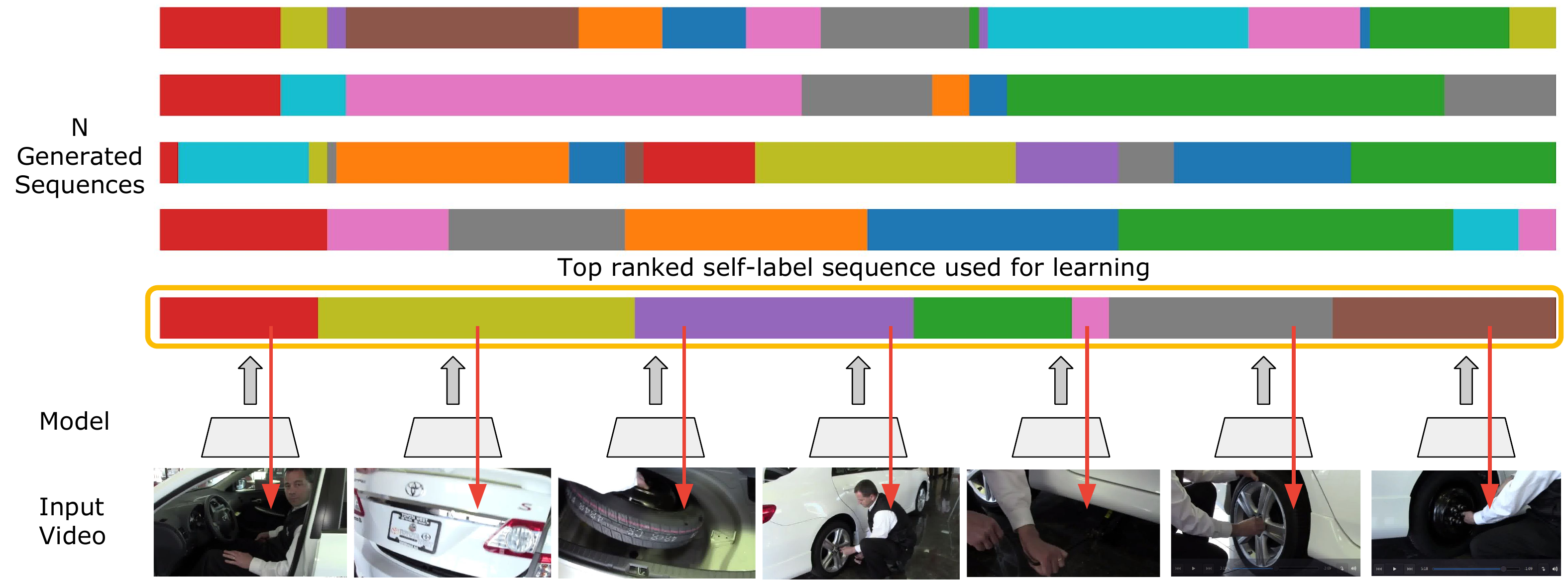}
\caption{Overview: Our model generates multiple sequences for each video which are ranked based on several constraints (colors represent different actions). The top ranked sequence is used as self-labels to train the action segmentation model. This processes is repeated until convergence. No annotations are used.}
    \label{fig:motivation}
\end{figure}

The central idea is to learn a stochastic model that generates multiple, different candidate sequences, which can be ranked based on instructional video constraints. The top ranked sequence is used as self-labels to train the action segmentation model. By iterating this process in an EM-like procedure, the model converges to a good segmentation of actions (Figure~\ref{fig:motivation}). In contrast to previous weakly~\cite{richard2017weakly,huang2016connectionist} and unsupervised~\cite{alayrac2016unsupervised,kukleva2019unsupervised} action learning works, our method only requires input videos, no further annotations are used.

\begin{figure}[t]
    \centering
    \includegraphics[width=0.75\linewidth]{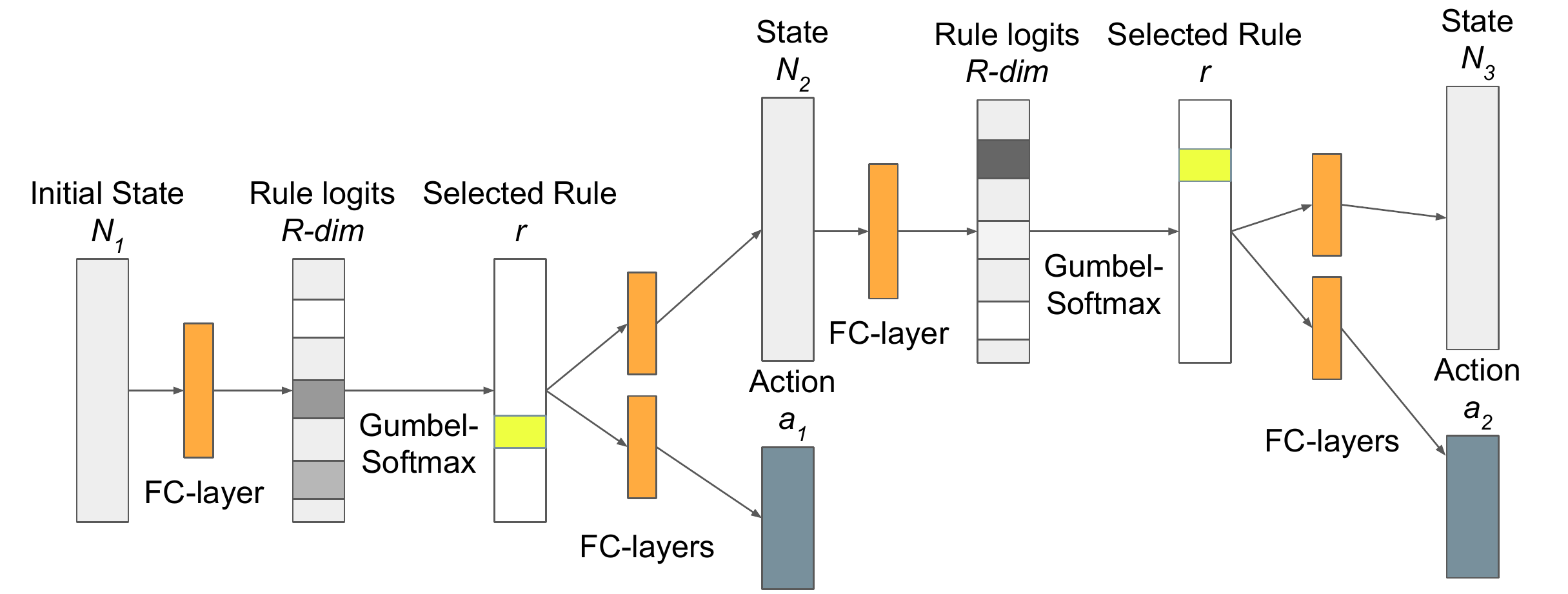}
    \caption{
    Overview of the stochastic recurrent model which generates an output action per step and a latent state (which will in turn generate next actions). Each time the model is run, a different rule is selected, thanks to the Gumbel-Softmax trick, leading to a different action and state. This results in multiple % different generated 
    sequences. % (see text for more details).
    %Overview of the stochastic recurrent model. Each time the Gumbel-Softmax is run, a different rule is selected, leading to a different action and state. This results in multiple % different generated 
    %sequences.
    }
    \label{fig:grammar-overview}
\vspace{-0.2cm}
\end{figure}

We evaluate the approach on multiple datasets and compare to previous methods on unsupervised action segmentation. We also compare to weakly-supervised and supervised baselines. Our unsupervised method outperforms all state-of-the-art models, in some cases considerably, with performance at times outperforming weakly-supervised methods. %We will \textbf{open source the code}.  

Our contributions are (1) a stochastic model capable of capturing multiple possible sequences, (2) a set of constraints and training method that is able to learn to segment actions without any labeled data.

%vspace{-2pt}
\textbf{Related Work}
%\vspace{-1pt}
Studying instructional videos has gained a lot of interest recently \cite{alayrac2016unsupervised,miech2019howto100m}, largely fueled by advancements in feature learning and  
activity recognition for videos. %~\cite{carreira2017quo,feichtenhofer2018slowfast}. 
However, most work on activity segmentation has focused on the fully-supervised case, which requires per-frame labels of the occurring activities.
Since it is expensive to fully annotate videos, weakly-supervised activity segmentation has been proposed. Initial works use movie scripts to obtain weak estimates of actions \cite{laptev2008learning} or  localize actions based on related web images \cite{gan2016webly}.  Several unsupervised methods have also been proposed~\cite{alayrac2016unsupervised,kukleva2019unsupervised,sener2018unsupervised}.  %Several datasets for learning from instructional videos %have been proposed. 
%have been introduced recently:
%Breakfast \cite{breakfast}, 50-salads \cite{salads}, Narrated Instructional Videos (NIV) \cite{alayrac2016unsupervised}, etc.

%\vspace{-2pt}
\section{Method}
%\vspace{-1pt}

% \subsection{Task}  % takes too much space, not really needed

Our goal is to discover atomic actions from a set of instructional videos, while capturing and modeling their temporal structure. Formally, given a set of videos $\mathcal{V} = \{V^1, V^2,...\}$ of a task or set of tasks, the objective is to learn a model that maps a sequence of frames $V^i = [I_t]_{t=1}^T$ from any video to a sequence of atomic action symbols $[a_t \in \mathcal{O}]_{t=1}^T$ where $\mathcal{O}$ is a set of possible action symbols.

%\begin{figure} [t]
%\begin{minipage}{.48\textwidth}
%    \centering
%    \includegraphics[width=0.8\linewidth]{figures/unsup-seg-full-model.pdf}
%    \caption{We use a CNN to process each frame, and concatenate those features with the state. Our sequential stochastic model processes each frame, generating a sequence of actions.}
%    \label{fig:full-model}
%\end{minipage}\hfill%
%\begin{minipage}{.48\textwidth}
%\vspace{-0.3cm}
%\end{figure} 

In the unsupervised case, similar to previous works \cite{alayrac2016unsupervised,kukleva2019unsupervised}, we assume no action labels or boundaries are given. Our model, however, works with a fixed $k$-the number of actions per task (analogous to setting $k$ in $k$-means clustering). This is not a very strict assumption as the number of expected atomic actions per instruction is roughly known.

\textbf{Sequential Stochastic Autoregressive Model.}
The model consists of three components: $(\mathcal{H}, \mathcal{O}, \mathcal{R})$ where $\mathcal{H}$ is a finite set of states, $\mathcal{O}$ is a finite set of output symbols, and $\mathcal{R}$ is a finite set of transition rules mapping from a state to an output symbol and next state. Importantly, this model is stochastic, i.e., each rule is additionally associated with a probability of being selected. To implement this, we use fully-connected layers and the Gumbel-Softmax trick \cite{jang2016categorical}.  The model is applied autoregressively to generate a sequence  (Figure~\ref{fig:grammar-overview}).
% commented out ,maddison2016concrete

For a video as input, we process each RGB frame by a CNN, resulting in a sequence of feature vectors. The model takes each feature as input and concatenates it with the state which is used as input to produce the output action. Once applied to every frame, this results in a sequence of actions.

 \begin{figure} 
    \centering
    \includegraphics[width=0.8\linewidth]{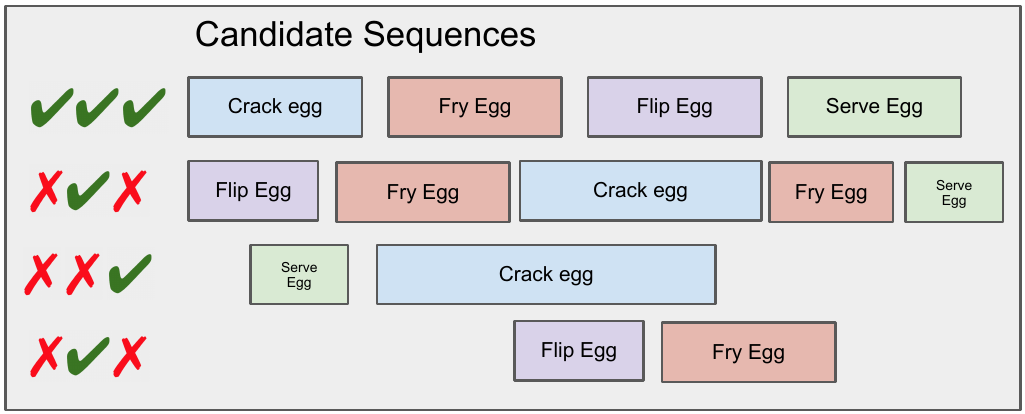}
    \caption{Multiple candidate sequences are generated and ranked. The best sequence according to the ranking function is chosen as the labels for the iteration.}
    \label{fig:candidates}   % \end{minipage}
\end{figure}

\textbf{Learning by Self-Labeling of Videos.}
In order to train the model without ground truth action sequences, we introduce an approach of learning by `self-labeling' videos. The idea is to optimize the model by generating self-supervisory labels that best satisfies the constraints required for atomic actions.
We first generate multiple candidate sequences, then rank them based on the instructional video constraints, which importantly require no labeled data.
Since the Gumbel-Softmax adds randomness to the model, the output can be different each time $G$ is run with the same input, which is key to the approach. The ranking function we propose to capture the structure of instructional videos has multiple components: (1) Every atomic action must occur once in the task. (2) Every atomic action should have similar lengths across videos of the same task. (3) Each symbol should reasonably match the provided visual feature.

\begin{figure}
    \centering
    \includegraphics[width=\linewidth]{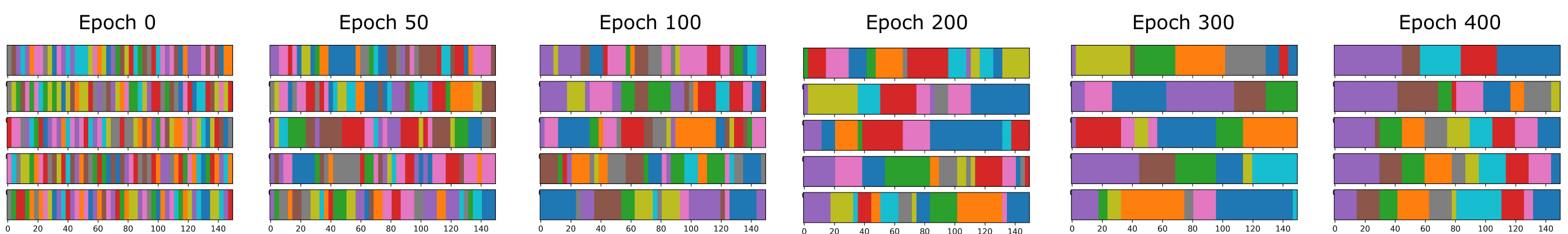}
    \caption{Candidate sequences at different stages of training. The sequences shown are the top 5 ranked sequences (rows) at the given epoch. %(columns) \[0, 50, 100, 200, 300, and 400\] training epochs. 
    The top one is selected as supervision for the given step. The colors represent the discovered action (with no labels).}
    \label{fig:example-generations}
\end{figure}

The best sequence according to the ranking is selected as the action labels for the iteration (Fig.~\ref{fig:candidates}), and the network is trained using a standard cross-entropy loss. We note that depending on the structure of the dataset, these constraints may be adjusted, or others more suitable ones can be designed. In Fig. \ref{fig:example-generations}, we show the top 5 candidate sequences and show how they improve over the learning process.

\textbf{Action Occurrence:}
Given a sequence $S$ of output actions, the first constraint ensures that every action appears once. Formally, it is implemented as $C_1(S) = |\mathcal{O}| - \sum_{a\in\mathcal{O}} \mathrm{App}(a)$, where $\mathrm{App}$ is 1 if $a$ is in $S$ otherwise 0.

\textbf{Modeling Action Length:}
This constraint ensures each atomic action has a similar duration across different videos. The simplest approach is to compute the difference in length compared to the average action length in the video.
We also compare to sampling the length from a distribution (e.g., Poisson or Gaussian). $C_2(S) = \sum_{a\in\mathcal{O}} (1-p(L(a, S)))$ where $L(a,S)$ computes the length of action $a$ in sequence $S$ and is Poisson or Gaussian. The Poisson and Gaussian distributions have parameters which control the expected length of the actions in videos. The parameters can be set statically or learned for each action.

\textbf{Modeling Action Probability:}
The third constraint is implemented using the separate classification layer of the network $p(a|f)$, which gives the probability of the frame being classified as action $a$. Formally, $C_3(S) = \sum_{t=1}^T (1-p(a_t|f_t))$, which is the probability that the given frame belongs to the selected action. This constraint is separate from the sequential model and captures independent appearance based probabilities.% We note that $a_t$ and $p_t$ are very similar, yet capture different aspects. $p_t$ is generated by a FC-layer applied independently to each frame, while $a_t$ is generated by the
auto-regressive model. We find that using both allows for the creation of the action probability term, which is useful empirically.

We can then compute the rank of any sequence as $C(S) = \gamma_1 C_1(S) + \gamma_2 C_2(S) + \gamma_3 C_3(S)$, where $\gamma_i$ weights the impact of each term. In practice setting $\gamma_2$ and $\gamma_3$ to $\frac{1}{|S|}$ and $\gamma_1=\frac{1}{|\mathcal{O}|}$ works well.

\textbf{Learning Actions:}
To choose the self-labeling, we sample $K$ sequences, compute the cost and select the one that minimizes the above cost function. This gives the best segmentation of actions (at this iteration of labeling) based on the defined constraints.

%\vspace{-2pt}
\paragraph{Cross-Video Matching}
%\vspace{-1pt}
The above constraints work well for a single video, however when we have multiple videos with the same actions, we can further improve the ranking function by adding a cross-video matching constraint. The motivation for this is that while breaking an egg can be visually different between two videos, the action is the same. 

Given a video segment the model labeled as an action $f_a$ from one video, a segment $\hat{f}_a$ the model labeled as the same action from a second video, and a segment $f_b$ the modeled labeled as a different action from any video, the cross-video similarity is computed using a triplet loss or a contrastive loss. As these functions are differentiable, they can be added to the loss or cost function or both.

%\vspace{-2pt}
\textbf{Self-labeling Training Method.}
%\vspace{-1pt}
We now describe the full training method, which follows an EM-like procedure. In the first step, we find the optimal set of action self-labels given the current model parameters and the ranking function. In the second step, we optimize the model parameters (and optionally some ranking function parameters) for the selected self-labeling. After taking both steps, we have completed one iteration. %For short, we call our full unsupervised method \textbf{\emph{UDA}} from \textbf{Unsupervised Discovery of Actions}. 

%\begin{savenotes*}
\setlength{\tabcolsep}{0.4em} % for the horizontal padding
\begin{table}
\centering
\small
   %\begin{minipage}{.31\textwidth}  % was 0.24
    \begin{tabular}{l|c|c|cc}
    \toprule
    Method & NIV (F1) & 50Sal (Acc) & BR (MoF) & BR (Jac) \\
    \midrule
    \multicolumn{2}{l}{\textbf{Supervised Baselines}}\\
    \midrule
     VGG from ~\cite{alayrac2016unsupervised} &  0.376 & 60.8 & 62.8 & 75.4 \\
    I3D & 0.472  & 72.8 &  67.8 & 79.4 \\
    AssembleNet \cite{Ryoo2020AssembleNet} & 0.558 & 77.6  & 72.5 & 82.1 \\
    \midrule
    \multicolumn{2}{l}{\textbf{Weakly-supervised}}\\
    \midrule
    %CTC (VGG) & 0.268   \\
    CTC \cite{huang2016connectionist} +~\cite{Ryoo2020AssembleNet} & 0.312 & 11.9 & 72.5 & 82.1 \\
    %CTC (Assemble) & 0.312  \\
    %ECTC (VGG) & 0.278  \\
    %ECTC (Assemble) & 0.334 \\
    HTK \cite{kuehne2017weakly} & - & 24.7 & - & -\\
    HMM + RNN \cite{richard2017weakly} & - & 45.5 & 33.3 & 47.3 \\
    NN-Viterbi \cite{richard2018neuralnetwork} & - & 49.4 & - & - \\
    ECTC \cite{huang2016connectionist} +~\cite{Ryoo2020AssembleNet}  & 0.334 & -  & 27.7 & - \\
    \midrule
    \multicolumn{2}{l}{\textbf{Unsupervised}}\\
    \midrule
    Uniform Sampling & 0.187 & - & - & - \\
    Alayrac et al. \cite{alayrac2016unsupervised} & 0.238 & -  & - & -\\
    Kukleva et al \cite{kukleva2019unsupervised} & 0.283 & 30.2 & 41.8 & -\\
    % GMM  & 0.265 \\
    % GreedySeq & 0.357 \\
    JointSeqFL, \cite{procel} & 0.373 & - & - & -\\
   %Ours (VGG) & 0.320 \\
    SCV \cite{li2020set} & - & - & 30.2 & - \\
    Sener et al. \cite{sener2018unsupervised} & - & - & 34.6 & 47.1 \\
    Ours & \textbf{0.457} & \textbf{39.7} & \textbf{43.5} & \textbf{54.4} \\
    \bottomrule
    \end{tabular}
%    \end{minipage}\hspace{5mm}%   % \hspace{10mm}
    \caption{Results on the NIV (left column), 50-salads (50Sal) (middle) and Breakfast(BR) (right) datasets. We report metrics adopted from prior work per each dataset, where available.}
    \label{tab:cvpr16-main-results}
\vspace{-0.1cm}  
\end{table}

\textbf{Segmenting a video at inference:} CNN features are computed for each frame and the learned model is applied on those features. During rule selection, we greedily select the most probable rule. Future work can improve this by considering multiple possible sequences (e.g., following the Viterbi algorithm).

%\vspace{-2pt}
\section{Experiments}
%\vspace{-1pt}
\footnotetext{For the weakly-supervised setting, we use activity order as supervision, equivalent to previous works.}

We evaluate our unsupervised atomic action discovery approach on multiple video segmentation datasets:
%, confirming that our self-generated action annotations form meaningful action segments.
%
%
%\noindent\textbf{Datasets:} 
%We compare results on 
(1) \textbf{50-salads dataset} \cite{salads}, which contains 50 videos of people making salads (i.e., a single task). The videos contain the same set of actions (e.g., cut lettuce, cut tomato, etc.), but the ordering of actions is different in each video, (2) \textbf{Narrated Instructional Videos (NIV) dataset} \cite{alayrac2016unsupervised}, which contains 5 different tasks (CPR, changing a tire, making coffee, jumping a car, re-potting a plant), %These videos have a more structured order.
(3) \textbf{Breakfast} \cite{breakfast} which contains videos of people making
breakfast dishes from various camera angles and environments.

\noindent\textbf{Evaluation Metrics:} 
%We follow the standard evaluation metrics adopted in each dataset. 
We follow all previously established protocols for evaluation in each dataset. We first use the Hungarian algorithm to map the predicted action symbols to action classes in the ground truth. Since different metrics are used for different datasets we report the previously adopted metrics per dataset.

\subsection{Comparison to the state-of-the-art}
We compare to previous state-of-the-art methods on the three datasets (Table \ref{tab:cvpr16-main-results}). Our approach provides better segmentation results than previous unsupervised approaches and even for some weakly-supervised methods.

\begin{figure}
    \centering
    \includegraphics[width=\linewidth]{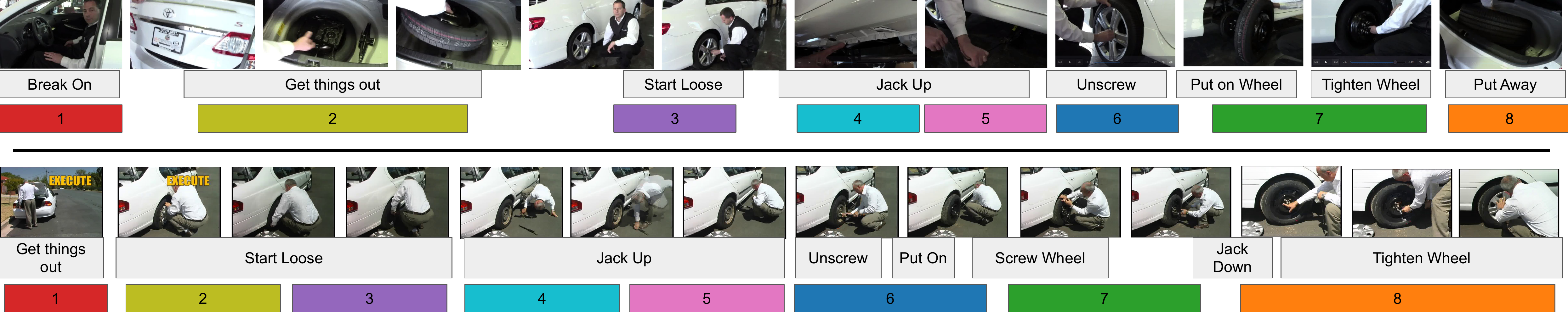}
    \caption{Two example videos from the `change tire' activity. The ground truth is shown in grey, the model's top rank segmentation is shown in colors. NIV dataset.}
    \label{fig:examples}
    \vspace{-5mm}
\end{figure}

\noindent\textbf{Qualitative Analysis} Fig. \ref{fig:example-generations} shows the generated candidate sequences at different stages of learning. It can be seen that initially the generated sequences are entirely random and over-segmented. As training progresses, the generated sequences start to match the constraints. After 400 epochs, the generated sequences show similar order and length constraints, and better match the ground truth (as shown in the evaluation).
Fig.~\ref{fig:examples} shows example results of our method.

%\vspace{-2pt}
\subsection{Ablation experiments}
%\vspace{-1pt}

\begin{table}[]
%    \end{minipage}\hfill%
    %
    %
%    \begin{minipage}{0.48\linewidth}
       \centering
       \small
    \begin{tabular}{l|cc}
    \toprule
    Cost & 50-Salads & Brkfst \\
    \midrule
    Randomly pick candidate & 12.5 & 10.8 \\
    No Gumbel-Softmax & 10.5 & 9.7 \\
    \midrule
    Occurrence ($C_1$)   & 22.4  & 19.8 \\
    Length ($C_2$)    & 19.6 & 17.8 \\
    $p(a|f)$ ($C_3$) & 21.5 & 18.8 \\
    \midrule
    $C_1 + C_2$ & 27.5 & 25.4 \\
    $C_1 + C_3$ & 30.3 & 28.4 \\
    $C_2 + C_3$ & 29.7 & 27.8 \\
    $C_1 + C_2 + C_3$ & 33.4 & 29.8 \\
    \bottomrule
    \end{tabular}
%    \end{minipage}
   \caption{Ablation with cost function terms$^2$}
    \label{tab:cost_fn}
 
\end{table}

\begin{table}[]
    \centering
    \footnotesize
    \begin{tabular}{l|ccccc|c}
    \toprule
 Method & chng  & CPR & repot  & make  & jump & Avg. \\
        &  tire &  & plant & coffee & car & \\
 \midrule
Alaryac et al. \cite{alayrac2016unsupervised}  & 0.41 & 0.32 & 0.18 & 0.20 & 0.08 & 0.238 \\
Kukleva et al. \cite{kukleva2019unsupervised} &-&-&-&-&-& 0.283\\
Ours VGG & 0.53 & 0.46 & 0.29 & 0.35 & 0.25 & 0.376 \\
Ours AssembleNet & 0.63 & 0.54 & 0.381 & 0.42 & 0.315 & 0.457 \\
\bottomrule
    \end{tabular}
    \caption{Comparison on the NIV dataset of the proposed approach on VGG and AssembleNet features.}
    \label{tab:features}
    \vspace{-3mm}
\end{table}

\begin{figure*}
    \centering
    %\makebox[\textwidth][c]{
    \includegraphics[width=0.19\linewidth,height=2cm]{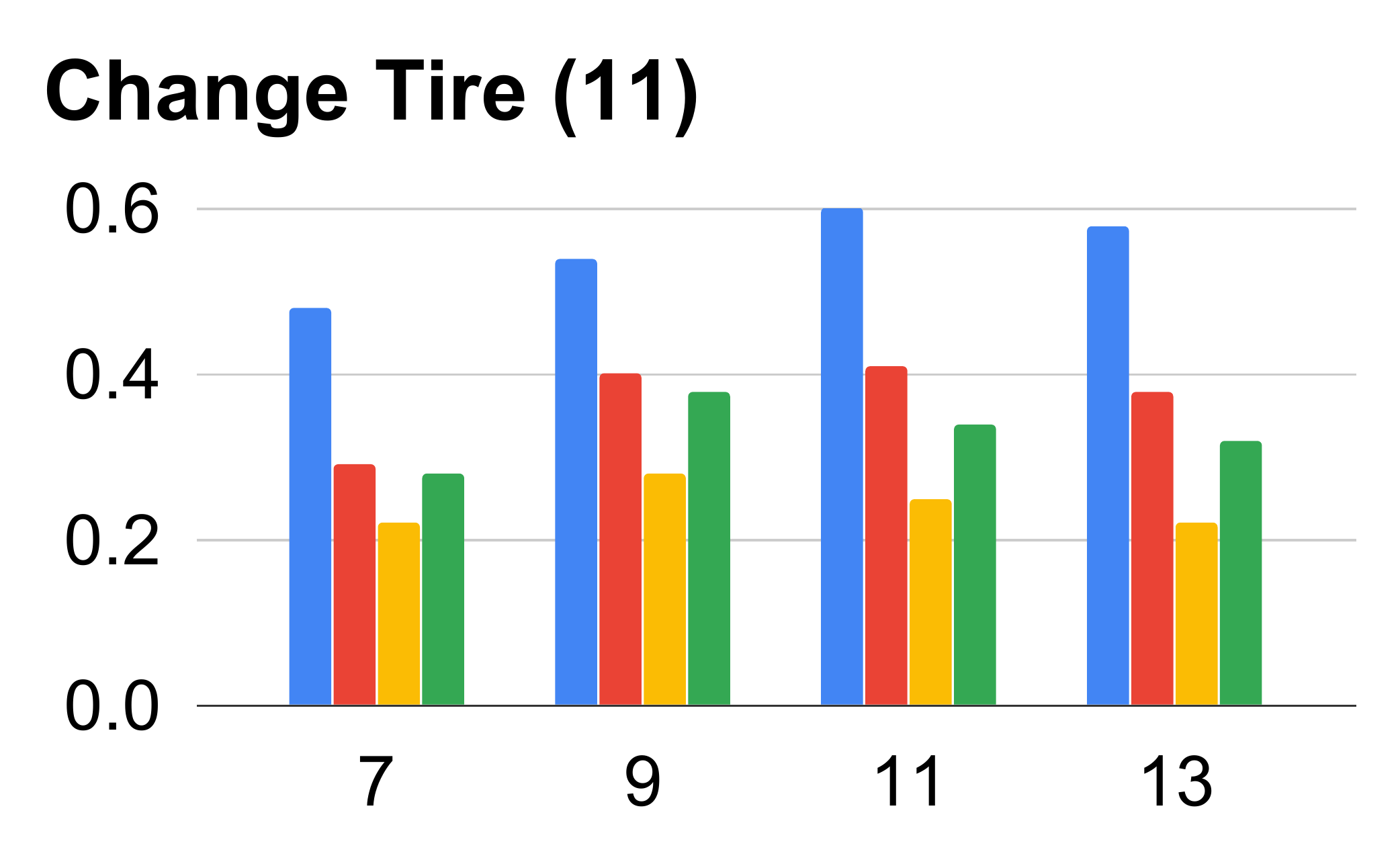}
    \includegraphics[width=0.19\linewidth, height=2cm]{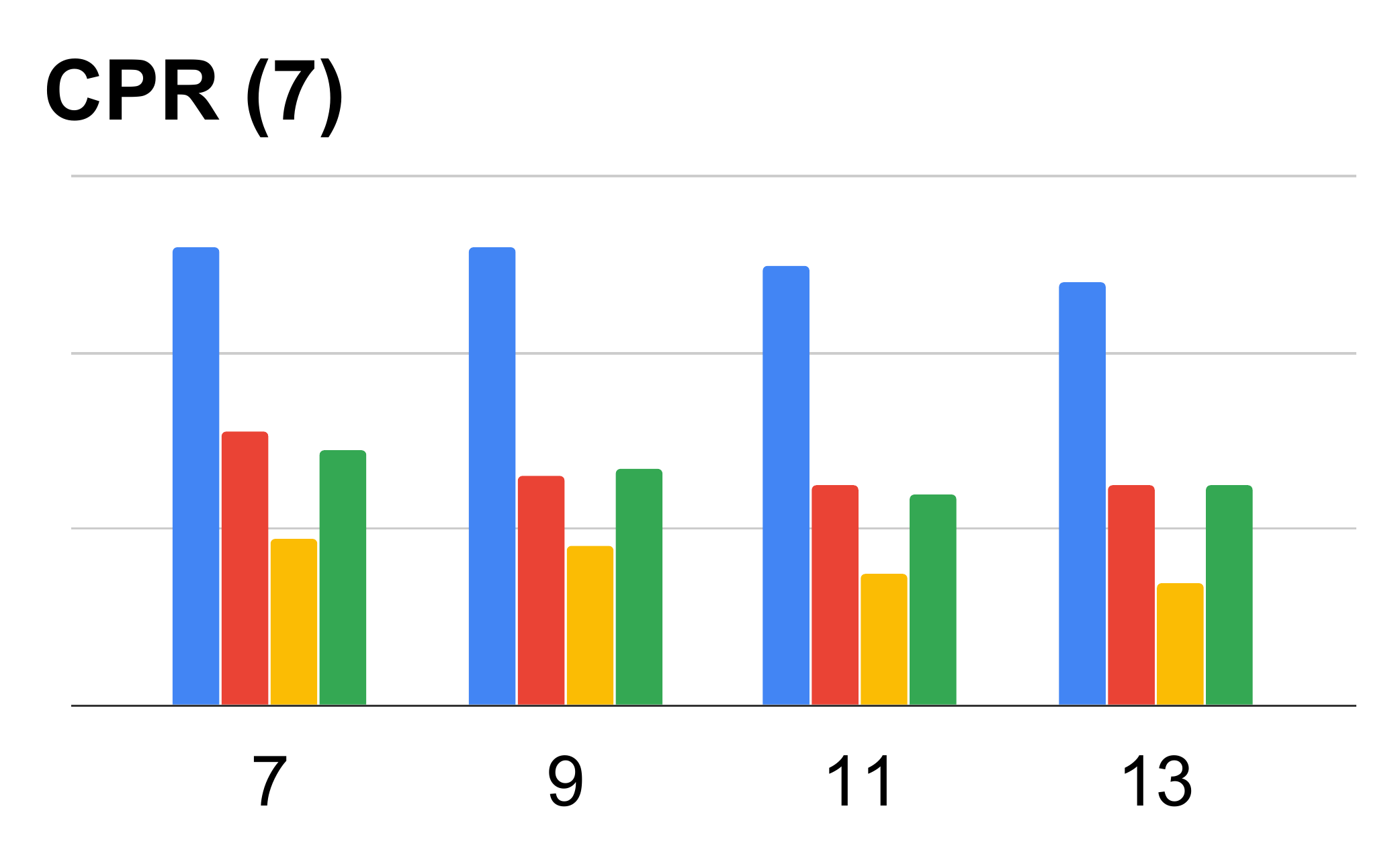}
    \includegraphics[width=0.19\linewidth, height=2cm]{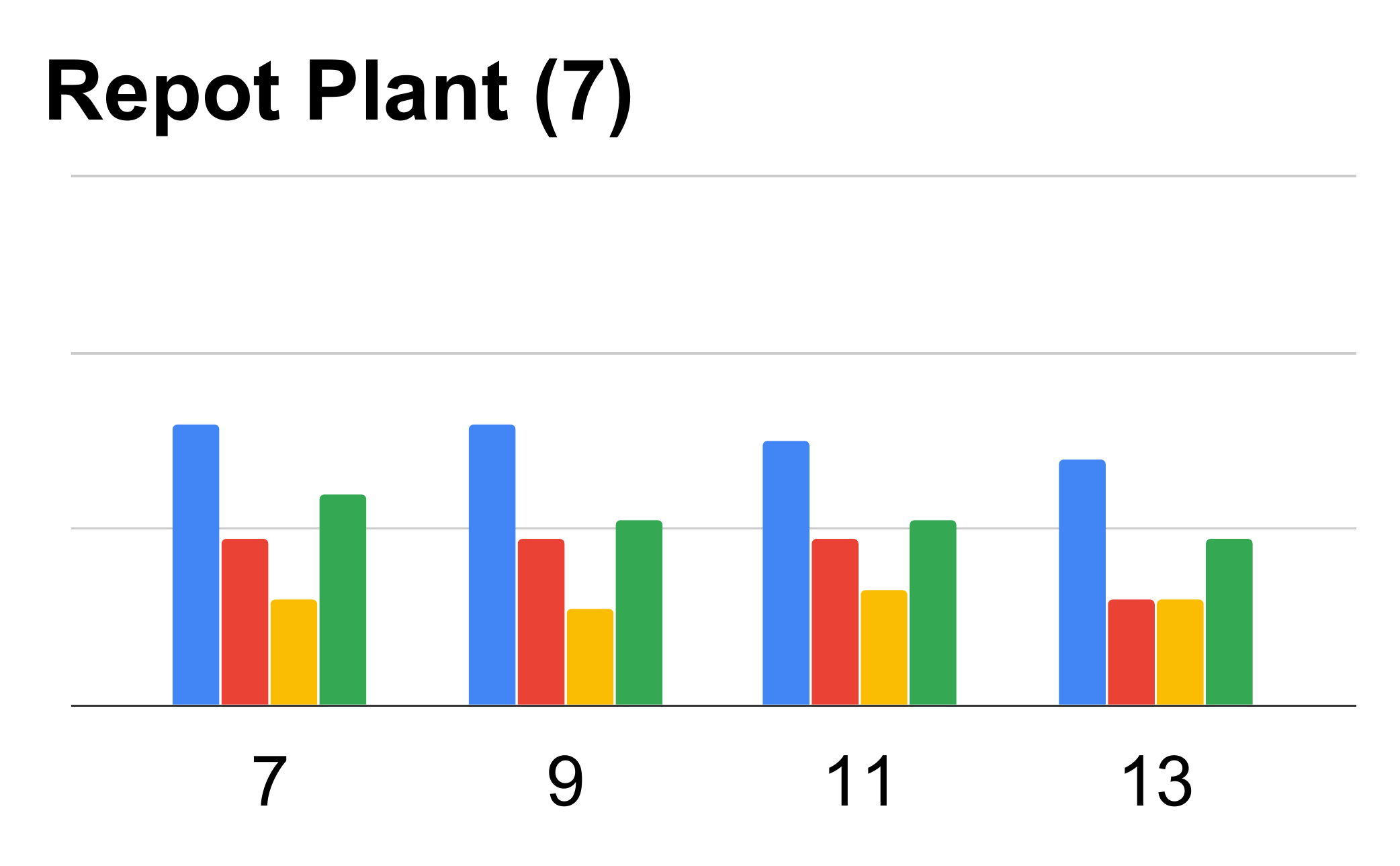}
    \includegraphics[width=0.19\linewidth, height=2cm]{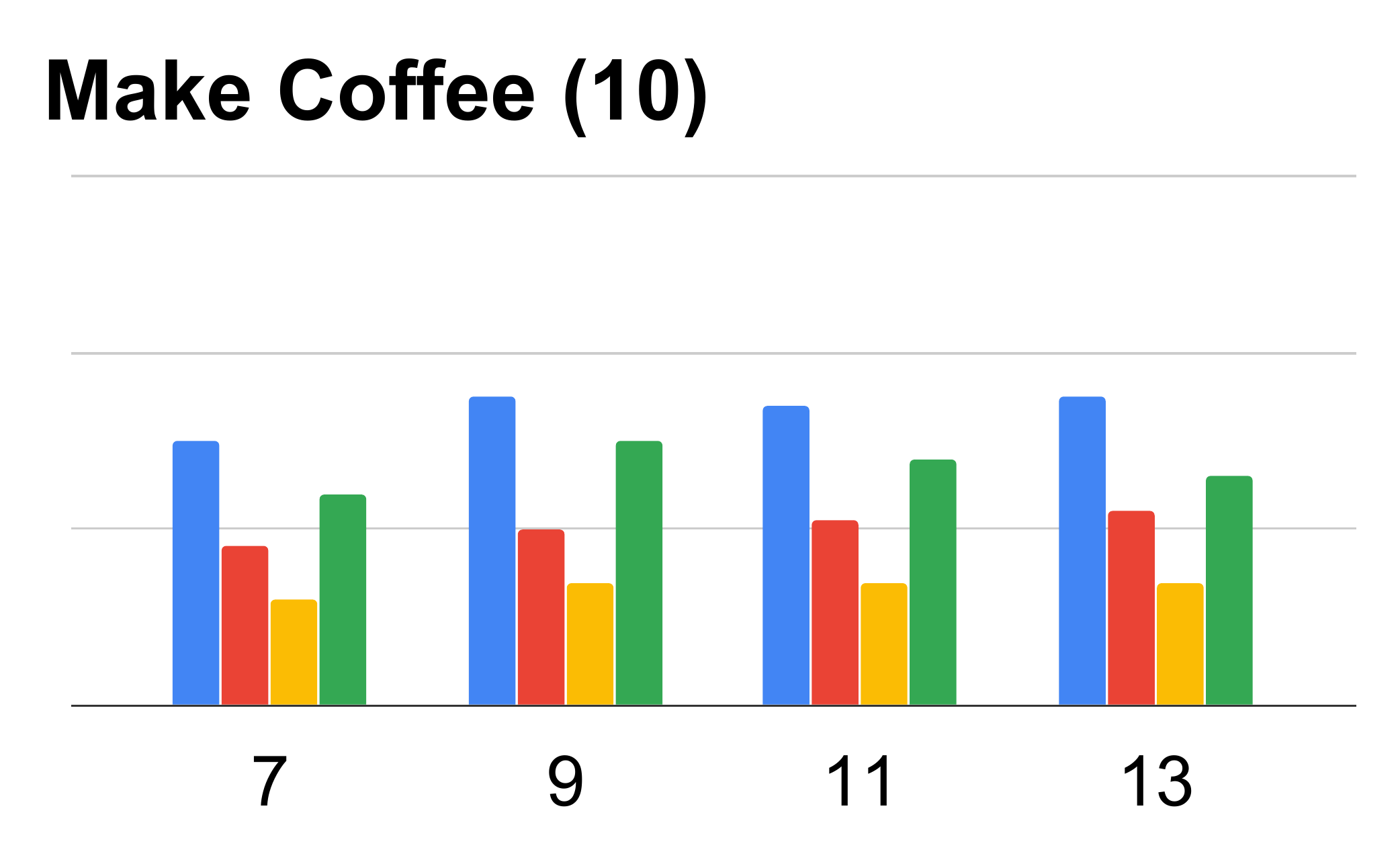}
    \includegraphics[width=0.19\linewidth, height=2cm]{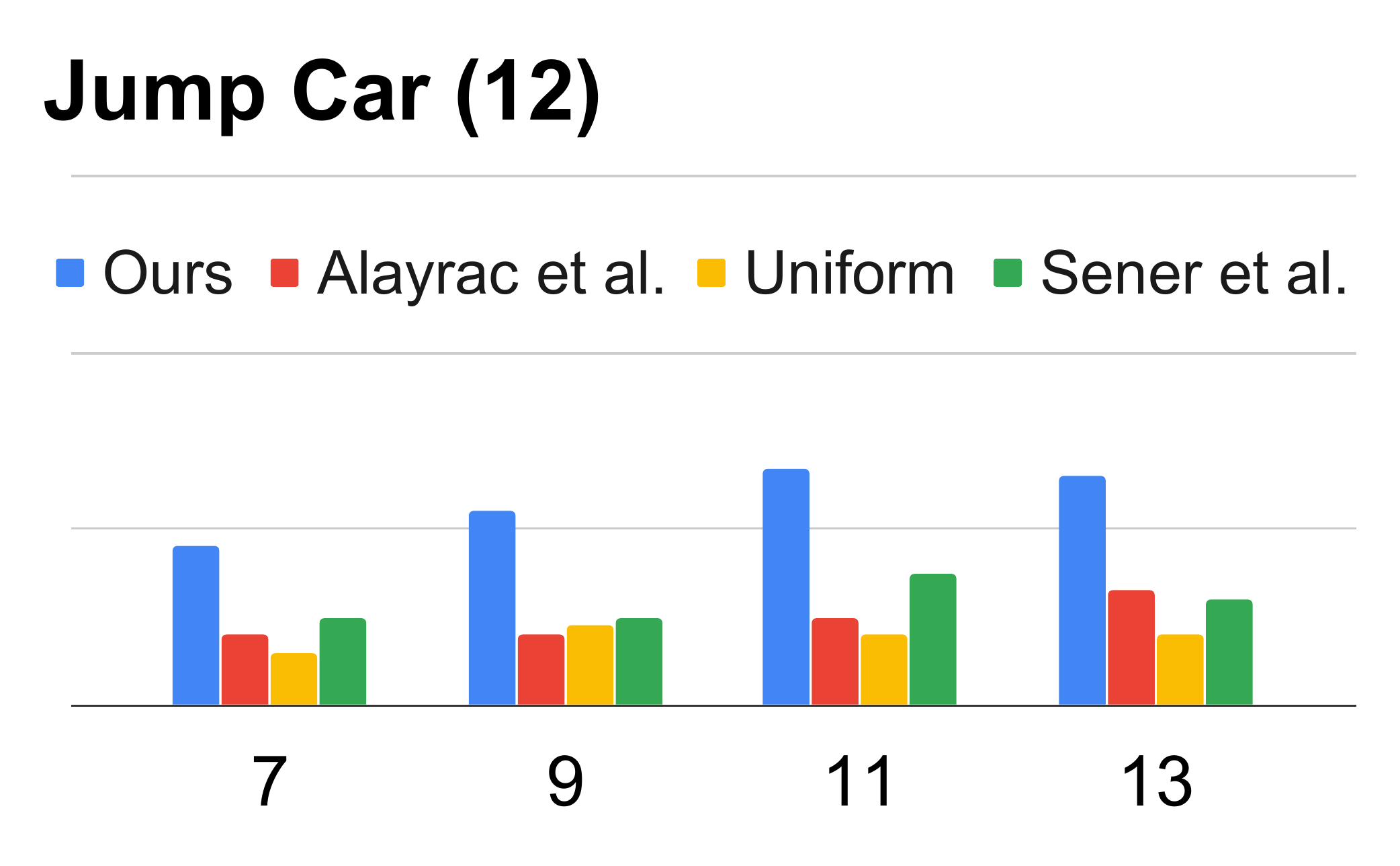}
    \caption{F1 value for varying the number of actions used in the model, compared to prior work. The number in parenthesis indicates the ground-truth number of actions for each activity. Full results are in the sup. materials.}
    \label{fig:num-steps}
    \vspace{-5mm}
\end{figure*}

\noindent\textbf{Effects of the cost function constraints.}
To determine how each cost function impacts the resulting performance, we compare various combinations of the terms. The results are shown in Table \ref{tab:cost_fn}. We find that each term is important to the self-labeling of the videos\footnote{These ablation methods do not use our full cross-video matching or action duration learning, thus the performances are slightly lower than the our best results.}. Generating better self-labels improves model performance, and each component is beneficial to the selection process. Intuitively, this makes sense, as the terms were picked based on prior knowledge about instructional videos. We also compare to random selection of the candidate labeling and a version without using the Gumbel-Softmax. Both alternatives perform poorly, confirming the benefit of the proposed approach.

\noindent\textbf{Varying the number of actions.}
As $\mathcal{O}$ is a hyper-parameter controlling the number of actions to segment the video into, we conduct experiments on NIV varying the number of actions/size of $\mathcal{O}$ to evaluate the effect this hyper-parameter has. The results are shown in Figure \ref{fig:num-steps}. Overall, we find that the model is not overly-sensitive to this hyper-parameter, but it does have some impact on the performance due to the fact that each action must appear at least once in the video.

\noindent\textbf{Features.}
As our work uses AssembleNet \cite{Ryoo2020AssembleNet} features, in Table \ref{tab:features} we compare the proposed approach to previous ones using both features. As shown, even when using VGG features, our approach outperforms previous methods.

{\small
\bibliographystyle{ieee_fullname}
\bibliography{bib}
}

\end{document}